\PassOptionsToPackage{margin=1in}{geometry}

\documentclass[11pt]{article}
\usepackage[final]{ACL2023}
\usepackage{times}
\usepackage{latexsym}
\usepackage[T1]{fontenc}
\usepackage[utf8]{inputenc}
\usepackage{microtype}
\usepackage{graphicx}
\usepackage{subcaption}
\usepackage{caption}
\usepackage[table]{xcolor}
\usepackage{multirow}
\usepackage{booktabs}
\usepackage[most]{tcolorbox}
\usepackage{tikz}
\usetikzlibrary{arrows.meta,decorations.pathmorphing,decorations.pathreplacing, calc}
\usepackage{hyperref}
\usepackage{capt-of}
\usepackage{float}
\usepackage{stfloats}
\usepackage{placeins} 
\usepackage{tabularx}
\usepackage{fontawesome5}

\newtcbox{\squared}[1][]{%
  on line, arc=2pt, boxrule=0.8pt,
  colback=green, colframe=white, 
  boxsep=1pt, left=1pt, right=1pt, top=1pt, bottom=1pt,
  colback=#1
}
\usepackage{listings}
\usepackage[scaled]{beramono}
\title{Language Re-generation: \\ An investigation into information locality effects on reconstruction}

\author{
 \textbf{Amirhossein Mohammadi},
 \textbf{Laurence E. Frank},
 \textbf{Albert Gatt},
 \textbf{Robert A. Bagheri}
\\
\\
 \textsuperscript{}Utrecht University,
\\
   \href{mailto:a.mohammadi@uu.nl}{a.mohammadi@uu.nl}
}

\begin{document}
    \maketitle
\begin{abstract}
Information locality, the tendency for syntactically related words to appear close together, shapes both human language processing and language model learning. While prior work has examined whether language models can acquire impossible languages, it remains unclear whether they can recover natural language from such input and what this reveals about their inductive biases. We address this by complementing learnability-based approaches with a reconstruction framework — fine-tuning GPT-2 models pre-trained on impossible languages to reconstruct natural English from three perturbation types. Our findings show that the recovered structures exhibit shorter dependency lengths than the original text, mirroring the locality preference observed in unconstrained language model generation and providing a quantitative signature of an architectural bias that learnability experiments alone do not reveal. Recovery difficulty increases with the degree of locality disruption. Structural recovery (dependency Triple F$_1$) dissociates from surface recovery (Exact Match), while fluency dissociates from faithful reconstruction under global shuffling. Sentence length further modulates performance: longer sentences facilitate recovery when local structure is preserved but lead to complete collapse under global shuffling. Finally, recovery difficulty tracks learnability difficulty across perturbation types, suggesting that information locality is the shared constraint governing both.

\begin{minipage}[t]{0.5\columnwidth}
\centering
\faGithub
\end{minipage}
\hspace{-1.8cm}
\href{https://github.com/amirhoseinMhmd/impossible_translation}{Repository}

\end{abstract} 
    \section{Introduction}

There are many conceivable linguistic systems, but not all langauges are equally learnable given human cognitive constraints~\cite{chomsky1957syntactic,Chomsky1986,Jackendoff2002,Pinker1994}. Natural languages share structural properties precisely because those properties keep them within cognitive reach \cite{Hawkins1994, Futrell2020Dependency}. One such property is \textit{Information locality}: the tendency for semantically and syntactically related elements to appear close together in linear order. This is rooted in memory limitations of the human brain \citep{Gibson2000Dependency, MussoEtAl2003BrocasArea}. \textit{Dependency Locality Theory} (DLT) formalizes this principle, predicting that processing difficulty increases as the linear distance between dependent elements grows, reflecting the limits of working memory during incremental comprehension \citep{Gibson1998Storage}. Cross-linguistic research corroborates the information locality principle through dependency length minimization (DLM), demonstrating that natural languages tend to converge on word order strategies that keep grammatically related elements adjacent or close \citep{Hawkins1994, Liu2008Dependency, Futrell2020Dependency, Temperley2018Dependency}.

\begin{figure}[t]
\centering
\footnotesize
\begin{tikzpicture}[
    x=0.9cm, y=0.9cm,
    inputblock/.style={
        rectangle, rounded corners=3pt,
        draw=red!50!black,
        fill=red!15,
        minimum width=2.6cm,
        minimum height=1cm,
        text width=2.5cm,
        align=center,
        font=\footnotesize
    },
    outputblock/.style={
        rectangle, rounded corners=3pt,
        draw=green!50!black,
        fill=green!15,
        minimum width=2.6cm,
        minimum height=1cm,
        text width=2.5cm,
        align=center,
        font=\footnotesize
    },
    llm_circle/.style={
        circle,
        draw=blue!70!black,
        fill=blue!20,
        minimum size=1.5cm,
        font=\bfseries\footnotesize,
        inner sep=2pt
    },
    flowarrow/.style={
        -{Stealth[length=2mm,width=1.3mm]},
        very thick,
    },
    zigarrow/.style={
        decorate,
        decoration={zigzag, segment length=5pt, amplitude=1.2pt},
        -{Stealth[length=2mm,width=2.5mm]},
        very thick,
    }
]

\node[font=\bfseries] at (0, 3.5) {Perturbed};
\node[font=\bfseries] at (0, 3) {sentence};

\node[font=\bfseries] at (6, 3.5) {Original};
\node[font=\bfseries] at (6, 3) {sentence};

\node[inputblock] (I1) at (0,  2) {The cat \squared[red!15, colframe=black]{R} mat the on sat};
\node[inputblock] (I2) at (0,  0) {The loudly barked dog};
\node[inputblock] (I3) at (0, -2) {Shire. Bag live Frodo the End at in};

\node[llm_circle] (LLM) at (3, 0) {LLMs};

\node[outputblock] (O1) at (6,  2) {The cat sat on the mat};
\node[outputblock] (O2) at (6,  0) {The dog barked loudly};
\node[outputblock] (O3) at (6, -2) {Frodo lives at Bag End in the Shire.};

\draw[zigarrow, draw=blue!70!black!40]      (I1.east) to[out=0,  in=140] (LLM.140);
\draw[zigarrow, draw=orange!80!black!50]   (I2.east) to[out=0,  in=180] (LLM.180);
\draw[zigarrow, draw=black!40]  (I3.east) to[out=0,  in=220] (LLM.220);

\draw[flowarrow, draw=blue!70!black!40] (LLM.40)  to[out=40,  in=180] (O1.west);
\draw[flowarrow, draw=orange!80!black!50]        (LLM.0)   to[out=0,   in=180] (O2.west);
\draw[flowarrow, draw=black!40]        (LLM.-40) to[out=-40, in=180] (O3.west);

\end{tikzpicture}

\caption{\footnotesize Translation task overview. Three separate LLMs are fine-tuned on specific perturbation types (top to bottom \textsc{PartialReverse}, \textsc{LocalShuffle}, \textsc{fullShuffle}). Each model translates its corresponding perturbed language input (left) into the original language output (right), recovering disrupted natural structure.}

\label{fig:paper-overview}
\end{figure}
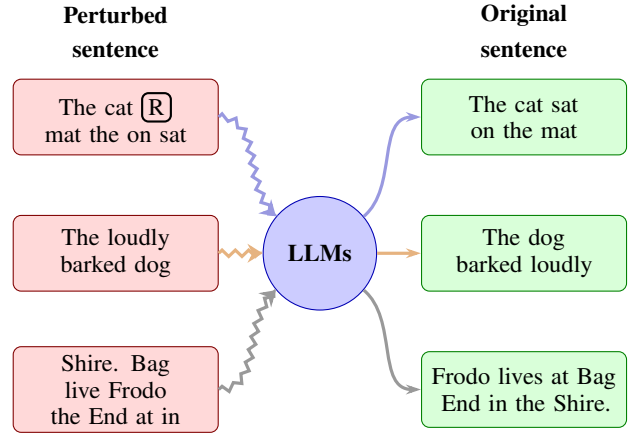

When dependent elements are separated from their heads by intervening and unrelated words, this order is disrupted, and the dependency structure of a sentence becomes harder to recover. 
As a result, the text becomes cognitively inaccessible even though its lexical content remains intact. Scrambling word order in violation of the syntactic constraints of a given language, through operations such as shuffling or reversal, increases dependency lengths and disrupts local syntactic relationships, producing inputs that are harder for humans to process despite intact vocabulary. Although some studies argue that LMs process such violations similarly to natural language \citep{ZivEtAl2025BiaslessLLMs}, \citet{kallini2024mission} demonstrate systematic differences in the learnability of languages which have undergone such perturbations, showing a clear inductive bias toward natural language structures. \citet{someya_information_2025} further establish that information locality is the primary driver of this bias. This bias extends to text generation: LMs exhibit a strong tendency to place semantically and syntactically related elements close together \citep{Khandelwal2018Sharp, Futrell2020Dependency}.

Despite this body of evidence, which suggests that LMs exhibit comparable tendencies to humans in terms of learnability, we still lack direct tests of the hypothesis that information is harder to recover as information locality is violated. In this paper, we address this question, by framing it as a translation task.
Translating perturbed inputs back into the original text directly tests whether LMs can identify scattered dependencies and restore their characteristic proximity. 
To address this, we take the {\bf pre-trained} GPT-2 models from \citet{kallini2024mission} as our starting point and fine-tune them to reconstruct valid English forms from inputs which undergo perturbations that disrupt locality to different degrees. We finetune one model per perturbation type. These models have already been exposed to impossible language distributions, so fine-tuning tests recovery directly without conflating the difficulty of acquisition with the difficulty of reconstruction. Investigating whether LMs can restore information locality from structurally degraded inputs provides theoretical insight into LM architectural biases \citep{kallini2024mission, Xu2025CanLM, Tran2018Importance, Khandelwal2018Sharp}, while practically informing how systems process corrupted or noisy text. We investigate two core questions: \textbf{First}, what does the pattern of recovery success and failure across perturbation types and dependency metrics reveal about LMs' inductive biases toward information locality? \textbf{Second}, does recovery difficulty scale with locality violation severity?

These experiments revealed three key findings. \textbf{First}, recovery performance varies systematically by perturbation type, directly reflecting the degree of locality violation. \textbf{Second}, dependency-based evaluation shows that LMs recover substantial syntactic structure even when the reconstructed string is not an exact match to the original, indicating that structural recovery is dissociable from surface-level reconstruction rather than tied to it; moreover, recovered structures systematically undershoot natural dependency length, revealing an architectural bias toward shorter-than-natural dependency arcs, consistent with DLM bias. \textbf{Third}, sentence length interacts with locality violation severity: longer sentences improve recovery for perturbations that preserve some local structure by providing richer contextual cues, but cause complete collapse for global shuffling, where dependency distances grow proportionally with sentence length. Together, these results suggest that LMs are sensitive to information-locality constraints but not fully bounded by them, exhibiting architectural biases that are partially aligned with human linguistic feasibility.
\footnote{All data and source code for this paper are available on \href{https://github.com/amirhoseinMhmd/impossible_translation}{GitHub}}

\definecolor{yellow}{HTML}{FFFFB0}    
\definecolor{peach}{HTML}{FFDAB9}     
\definecolor{red}{HTML}{FFB9B9}       
\definecolor{orange}{HTML}{FFD2AA}    
\definecolor{beige}{HTML}{F5F5D7}     
\definecolor{blue}{HTML}{B4DCFF}      
\definecolor{green}{HTML}{C3FFC3}     
\definecolor{purple}{HTML}{E1C8FF}    
\definecolor{pink}{HTML}{FFC8DC}      
\definecolor{mint}{HTML}{BEFFD2}      
\definecolor{coral}{HTML}{FFCDCB}     
\definecolor{lavender}{HTML}{C8C3FF}  
\definecolor{cyan}{HTML}{BEEFF5}      
\definecolor{mauve}{HTML}{EBCFFF}     
\definecolor{teal}{HTML}{B4EBE1}      

\definecolor{ignoredred}{HTML}{DC0000}
\definecolor{learnedteal}{HTML}{2E9F54}
\definecolor{softyellow}{HTML}{FFFEF7}
\begin{table*}[ht]
\centering
\small
\setlength{\tabcolsep}{4pt}
\begin{tabular}{@{} l| p{4cm} |p{7.5cm} @{}}
\toprule
\textbf{Language} & \textbf{Example 1} & \textbf{Example 2} \\
\midrule
\textsc{Original Text}
  & \squared[red]{It}\squared[blue]{is}\squared[yellow]{nice}\squared[purple]{in}\squared[orange]{there}
  & \squared[yellow]{we}\squared[teal]{'d}\squared[pink]{need}\squared[beige]{to}\squared[mauve]{look}\squared[lavender]{at}\squared[orange]{it}\squared[coral]{again},\squared[peach]{would}\squared[red]{n't}\squared[cyan]{we}\\
  \midrule
\textsc{LocalShuffle(w=3)}
  & \squared[yellow]{nice}\squared[blue]{is}\squared[red]{It}\squared[orange]{there}\squared[purple]{in}
  & \squared[yellow]{we}\squared[teal]{'d}\squared[pink]{need}\squared[beige]{to}\squared[orange]{it}\squared[mauve]{look}\squared[coral]{again}\squared[lavender]{at},\squared[peach]{would}\squared[red]{n't}\squared[cyan]{we}\\
\midrule
\textsc{PartialReverse}
  & \squared[red]{It}\squared[blue]{is}\squared[green, colframe=black]{R}\squared[orange]{there}\squared[purple]{in}\squared[yellow]{nice}
  &\squared[yellow]{we}\squared[teal]{'d}\squared[pink]{need}\squared[green, colframe=black]{R}\squared[cyan]{we}\squared[red]{n't}\squared[peach]{would},\squared[coral]{again}\squared[orange]{it}\squared[lavender]{at}\squared[mauve]{look}\squared[beige]{to}\\

\midrule

\textsc{FullShuffle}
  & \squared[yellow]{nice}\squared[blue]{is}\squared[orange]{there}\squared[red]{It}\squared[purple]{in}
  &\squared[coral]{again},\squared[lavender]{at}\squared[red]{n't}\squared[yellow]{we}\squared[orange]{it}\squared[peach]{would}\squared[cyan]{we}\squared[pink]{need}\squared[mauve]{look}\squared[teal]{'d}\squared[beige]{to}\\

\bottomrule
\end{tabular}
\caption{Example data for each perturbation function. For ease of comparison, we use a similar visualisation to \citet{kallini2024mission}.}
\label{tab:data_sampla}
\end{table*}

\section{Background}

\paragraph{Cognitive constraints on language learning}

One goal of linguistics is to distinguish natural languages from the vast space of conceivable grammars — to identify which structural properties define human language \citep{Saussure1916, Hockett1958, Lyons1968}. Chomsky's Universal Grammar (UG) framework formalizes this distinction, proposing that all human languages share constrained structural principles derived from innate cognitive mechanisms \citep{chomsky1965aspects}. Crucially, a language can be formally describable without being cognitively learnable. In other words, the question of whether a grammatical formalism can represent a linguistic system is distinct from the question of whether humans can acquire that system  \citep{Moro2016Impossible, MussoEtAl2003BrocasArea}. Under thiz view,   `impossible languages' are those whose grammars humans cannot acquire despite exposure, not because they cannot be formally characterised, but because they violate cognitive constraints \citep{Moro2016Impossible}.

\noindent
Empirical evidence for this boundary comes from acquisition studies and neurolinguistics. \citet{Smith1993} showed that humans successfully acquire artificial languages consistent with known principles governing natural languages, but fail to learn those that violate them, suggesting certain grammars remain unlearnable regardless of exposure. Neurolinguistic studies corroborate this: brain regions for syntactic processing selectively respond to natural language-compatible structures \citep{Moro2016Impossible, MussoEtAl2003BrocasArea}. \citet{Nefdt2024} further argues that identifying possible grammar boundaries requires grounding in theoretical linguistic competence, not solely distributional regularities. Yet this evidence remains limited — \citet{Smith1993} focused on a single individual and most theoretical claims lack systematic empirical testing — leaving the boundary between possible and impossible languages underspecified.

\noindent
Among the cognitive constraints that define this boundary, information locality is the most robustly documented. Cross-linguistic research demonstrates that natural languages systematically minimize dependency length to keep grammatically related elements adjacent \citep{Gibson2000Dependency, Gibson1998Storage, Hawkins1994, Liu2008Dependency, Futrell2020Dependency, Temperley2018Dependency}. \citet{someya_information_2025} establishes information locality as the primary driver of learnability differences between possible and impossible languages. This makes information locality not merely a descriptive property of natural language but a diagnostic for the possible/impossible boundary itself. \citet{mollica_composition_2020} find evidence that the human brain can recover meaningful dependencies from natural language strings even if they violate word order constraints, as long as dependents are located in close proximity.

\paragraph{LMs as models of linguistic competence}

Standard neural architectures do not automatically respect the constraints that define possible languages. \citet{hahntheoretical2020} shows that self-attention architectures have theoretical limitations when syntactic dependencies span arbitrary distances, and \citet{mitchell2020priorless} demonstrate that recurrent neural networks learn highly unnatural patterns that humans do not acquire — highlighting a potential misalignment between human and neural inductive biases.

\noindent
The extent of this misalignment, however, is not uniform across all linguistic phenomena. \citet{abdou2022word} show that models trained on shuffled text retain non-trivial word order information; \citet{huang2023lexinvariant} demonstrate that lexically invariant transformer variants match standard models under long contexts; and \citet{ebrahimi2020self} show that self-attention networks learn certain hierarchical languages with appropriate cues. However, \citet{deletang2023neural} argue that standard architectures face considerable challenges in representing complex hierarchical dependencies without explicit structured memory.

\noindent
\citet{kallini2024mission} demonstrate that LLMs exhibit systematic learnability differences between possible and impossible languages, showing biases toward natural patterns while struggling with highly unnatural systems. \citet{Xu2025CanLM} find that GPT-2 shows learning slowdowns aligned with typological universals when acquiring implausible counterfactual languages, though it can ultimately learn them. Together, these findings show that LLM biases only partially overlap with human cognitive constraints — leaving open whether LLMs can actively recover possible language from impossible inputs by reversing the locality violations that define the boundary.

    \section{Method}
The main goal of this research is to train an LLM to recover meaningful strings from perturbed inputs, where the latter can loosely be regarded as strings of an `impossible language'. Formally, we define a perturbation function $f: A \rightarrow A'$ where $A$ represents standard English text and $A'$ is the perturbed version.
Our objective is to train a model $\mathcal{M}$ such that $\mathcal{M}(A'):= A$, effectively reversing the perturbation and recovering the original linguistic information.

A critical constraint is that $\mathcal{M}$ must not have been pre-trained on $A$. If $\mathcal{M}$ had prior exposure to $A$ (e.g. standard English), it could leverage that knowledge rather than genuinely learning to recover linguistic structure from perturbed input, confounding our evaluation of information recovery. 

To satisfy this requirement, we adopt \textbf{the models pre-trained by} \citet{kallini2024mission} as the starting point for our fine-tuning process, since these models were trained from scratch on impossible languages, generated by applying different perturbation functions to standard English sentences.

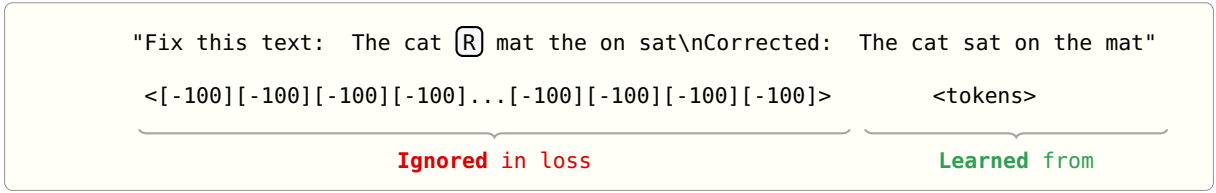
\begin{figure*}[!t]
\centering
\footnotesize
\begin{tcolorbox}[
  colback=softyellow,
  colframe=black!40,
  boxrule=0.4pt,
  width=\textwidth,       
]
\ttfamily\footnotesize
\hspace{1cm} "Fix this text: The cat \squared[gray!70, colframe=black,colback=black!3]{R} mat the on sat\textbackslash nCorrected: The cat sat on the mat"\\[10pt]
\hspace*{1cm} \texttt{ <[-100][-100][-100][-100]...[-100][-100][-100][-100]>}
\hspace{1cm} \texttt{<tokens>}
\\
\vspace{.5cm}
\begin{tikzpicture}[overlay, remember picture]
  \draw[gray!70, thick, decorate, decoration={brace, amplitude=3pt,mirror}]
    (1.1,0) -- (10.5,0)
    node[midway, below=5pt, font=\small, text=ignoredred]{\textbf{Ignored} in loss};
  \draw[gray!70, thick, decorate, decoration={brace, amplitude=3pt,mirror}]
    (10.7,0) -- (14.7,0)
    node[midway, below=5pt, font=\small, text=learnedteal]
      {\textbf{Learned} from};;
\end{tikzpicture}
\end{tcolorbox}
\caption{\footnotesize{The mechanism of masking labels and calculating loss}}
\label{fig:masking}
\end{figure*}

\subsection{Data Generation}
We use the \textsl{bnc\_spoken} and \textsl{Gutenberg} subsets of BabyLM \citep{WarstadtEtAl2023BabyLM}. These subsets allow us to analyse the impact of contrasting sentence lengths on translation performance. Applying perturbations to these subsets yields paired examples (perturbed, original), which we use for fine-tuning.
Following \citet{kallini2024mission}, we apply three perturbation functions (Table~\ref{tab:data_sampla}).

\textbf{\textsc{LocalShuffle:}} shuffling tokens randomly within a local window (size 3 in this study), disrupting the sequential arrangement while maintaining words within bounded distances. 

\textbf{\textsc{FullShuffle:}} shuffling tokens globally at random, disrupting the sequential arrangement throughout the entire text, thereby also disrupting information locality.

\textbf{\textsc{PartialReverse:}} A random starting point is selected within the sentence, and an \squared[green, colframe=black]{R} token is placed in this position, and subsequent tokens are reversed in order. This creates a partition where the initial segment remains unchanged, the final segment is reversed, and \squared[green, colframe=black]{R} marks the boundary.

These three perturbations from \citet{kallini2024mission} span a spectrum of locality violation: \textsc{LocalShuffle} disrupts local order within bounded windows, \textsc{PartialReverse} disrupts order from a random point, preserving the initial segment, and \textsc{FullShuffle} produces the most severe violation by scattering tokens globally, making dependency recovery maximally difficult.

We created three datasets, each containing paired text (impossible, possible), by applying one perturbation strategy to each. To investigate training scale effects, we use datasets of $100K$ sample pairs from \textsl{bnc\_spoken}.

\subsection{Model Fine-tuning}\label{sec:model-finetuning}

The models were fine-tuned by combining the causal language modelling paradigm of GPT-2 \citep{RadfordEtAl2019GPT2} with a masked label strategy. Each training sample consists of a pair: a perturbed text and its corresponding original version. These pairs are formatted into the following instruction-response structure:

\begin{tcolorbox}[
  colback=softyellow,
  colframe=black!40,
  boxrule=0.4pt,
  arc=2pt,
  width=\columnwidth,
  left=4mm, right=4mm, top=2mm, bottom=2mm,
]
\ttfamily\footnotesize
Fix this text: <impossible\_text>
\\
Corrected: <possible\_text><|endoftext|>
\end{tcolorbox}

To ensure the model learns to generate corrections rather than memorising instructions, masked labelling is applied. The instruction tokens (sequence start to \texttt{Corrected}) receive label value $[-100]$, while the response tokens (correct text and \texttt{<|endoftext|>} token) retain their IDs as labels. By excluding the input from the loss calculation, the model is still presented with the full input during the forward pass, but the loss is computed only over the target (correct) tokens \citep{ouyang2022training}. The GPT-2 tokenizer was extended with special tokens \{\squared[green, colframe=black]{R}\} before fine-tuning to accommodate perturbation markers. Figure \ref{fig:masking} displays the label masking and loss calculation.

\subsection{Evaluation metrics}\label{metrcis}

To assess how well the model recovers syntactic structure from perturbed input, we employ Exact Match (EM), measuring perfect reconstruction~\citep{rajpurkar-etal-2016-squad}; and the BLEU score quantifying n-gram overlap~\citep{papineni-etal-2002-bleu}. We use BLEU purely as a measure of the extent to which a model is able to recover surface sequences of varying length in the original. We also consider two complementary metrics based on dependency triples:

\paragraph{Average Dependency Length (ADL).}
We compute the mean absolute positional distance between each non-root token and its syntactic head, based on head-dependent relationships identified in the {\em original} sentence:
$\text{ADL} = \frac{1}{n} \sum_{i=1}^{n} | \text{pos}(w_i) - \text{pos}(\text{head}(w_i)) |$,
where $n$ is the number of non-root tokens. Scrambled text disrupts this locality, producing higher ADL.

\paragraph{Dependency Triple F1.}
From each parse, we extract triples $(w_d, r, w_h)$ where $w_d$ is the dependent, $r$ is the relation label, and $w_h$ is the head. We compute set-based precision, recall, and F1 between the triple sets in the model's prediction and the original (unperturbed) string. A triple matches only if all three elements are identical. As a baseline, we also compare the scrambled input against the actual text to quantify how many arcs survive perturbation by chance \citep{buchholz2006conll, cai-knight-2013-smatch}.

For both ADL and Dependency Triple F1, dependency parses were produced using the \texttt{en\_core\_web\_trf} model in spaCy.

\subsection{Experimental Design}

We conduct three experiments testing whether reconstruction difficulty scales with locality violation severity. We hypothesise that perturbations more severely disrupting the proximity of related elements should be harder to reverse, reflecting a gradient rather than a categorical boundary. If LLMs exhibit locality constraints, translation performance should degrade as violations intensify: longer sentences increase difficulty by expanding distances between related elements and enlarging reordering search spaces, while perturbation types should vary by locality disruption severity. For full fine-tuning details see  \ref{sec:finetune_details}. 

\paragraph{Experiment 1: Effect of training sample size.} This experiment investigates how training data size influences translation performance. Models were fine-tuned on the $100K$ subset of \textsl{bnc\_spoken}, with learning progress evaluated at multiple checkpoints using our four metrics (\S~\ref{metrcis}).
This multi-metric approach suits our task because perturbed and original texts share vocabulary but differ in token order, making both strict reconstruction (EM) and partial accuracy (BLEU, Triple F1), and structural locality (ADL)  informative. The hypothesis is that larger datasets will improve performance across all perturbations and metrics, but with different learning rates: \textsc{PartialReverse} should be learned most readily as it preserves the initial segment of each sentence, while \textsc{FullShuffle} will require the most examples as global shuffling completely destroys all dependency structure, with \textsc{LocalShuffle} showing intermediate difficulty.

\begin{table}[!tbp]
\centering
\footnotesize
\begin{tabular}{lccc}
\hline
\textbf{Dataset} & &\textbf{Size} & \textbf{Average Length}\\
\hline
\textsl{bnc\_spoken} & train & $100K$ & $12.77$ \\ 
\textsl{bnc\_spoken} & test & $1K$ & $11.72$ \\ 
\hline
\textsl{Gutenberg} & train & $100K$ & $40.61$ \\
\textsl{Gutenberg} & test & $1K$ & $46.78$ \\
\hline
\end{tabular}
\caption{\footnotesize{Average sentence lengths for the training and test datasets.}}
\label{tab:dataset-avg-len}
\end{table}

\paragraph{Experiment 2: Effect of sentence length.} Building on the locality principle introduced above, this experiment tests how sentence length affects recovery difficulty. Results were compared between the \textsl{bnc\_spoken} dataset and the \textsl{Gutenberg} dataset (see Table \ref{tab:dataset-avg-len}). The hypothesis is that longer sentences provide more training signal, potentially improving overall performance, but simultaneously increase difficulty for perturbations disrupting information locality. Specifically, \textsc{LocalShuffle} and \textsc{PartialReverse} are expected to show complex effects: benefits from additional training content may be offset by challenges from related words scattered over greater distances, which exponentially expand the space of possible reorderings and integration costs. \textsc{FullShuffle} is expected to show the greatest difficulty with longer sentences, as global shuffling exponentially expands the reordering search space and provides no preserved structure to guide reconstruction.

\begin{figure}[!t]
    \centering
    \begin{subfigure}{0.49\linewidth}
          \includegraphics[width=\linewidth]{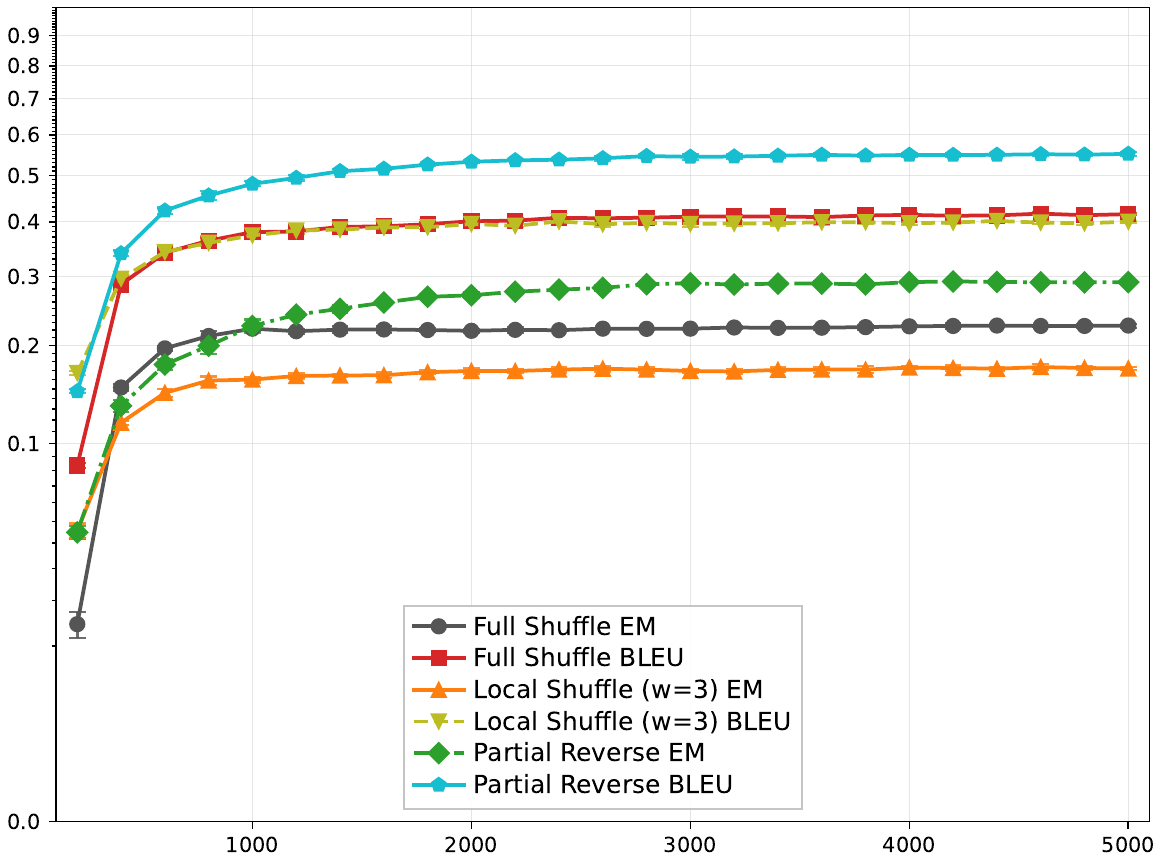}
          \caption{\textsl{bnc\_spoken}}
          \label{fig:bnc:em-bleu}
    \end{subfigure}
    \begin{subfigure}{0.49\linewidth}
        \includegraphics[width=\linewidth]{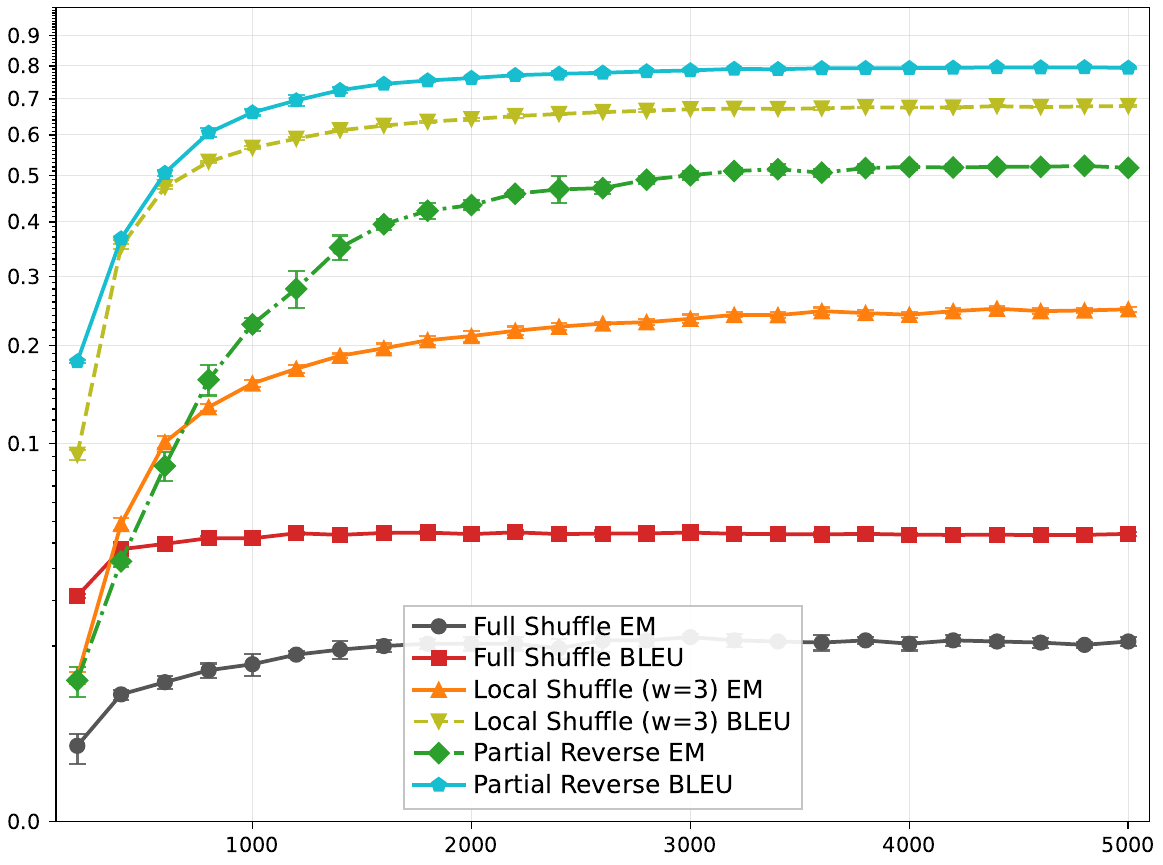}
        \caption{\textsl{Gutenberg}}
        \label{fig:gutenberg:em-bleu}
    \end{subfigure}
\caption{\footnotesize{Training curves comparing short (\subref{fig:bnc:em-bleu}) versus long (\subref{fig:gutenberg:em-bleu}) sentences. Longer contexts improve all perturbations, with \textsc{PartialReverse} showing largest gains.}
}
\label{fig:em-bleu}
\end{figure}

\paragraph{Experiment 3: Quality of generated text.} Because the goal of this research is to translate disturbed languages and recover information, this experiment evaluates whether generated texts converge toward natural language distributions using perplexity. We measure against GPT-2 base on standard English to measure how well generated text aligns with natural language distributions \citep{JelinekEtAl1977Perplexity}. Lower perplexity values indicate greater consistency with natural language distributions \citep{RadfordEtAl2019GPT2}. 
We hypothesise that prediction perplexity will converge toward that of the original text as training progresses, with rates varying by perturbation type.  Specifically, \textsc{FullShuffle} and \textsc{LocalShuffle} inputs are expected to show the highest initial perplexity given their severe locality disruption, with predictions converging toward natural language distributions as training progresses. \textsc{PartialReverse} inputs should show lower initial perplexity due to partial structure preservation, with predictions converging more rapidly toward baseline.

To ensure reproducibility, all fine-tunings were repeated across five different random seeds. The figures report average performance across all seeds, with error bars showing standard deviation.

    \section{Experimental results}
\begin{figure*}[tbp]
\begin{subfigure}{0.5\columnwidth}
  \centering
  \includegraphics[width=\linewidth]{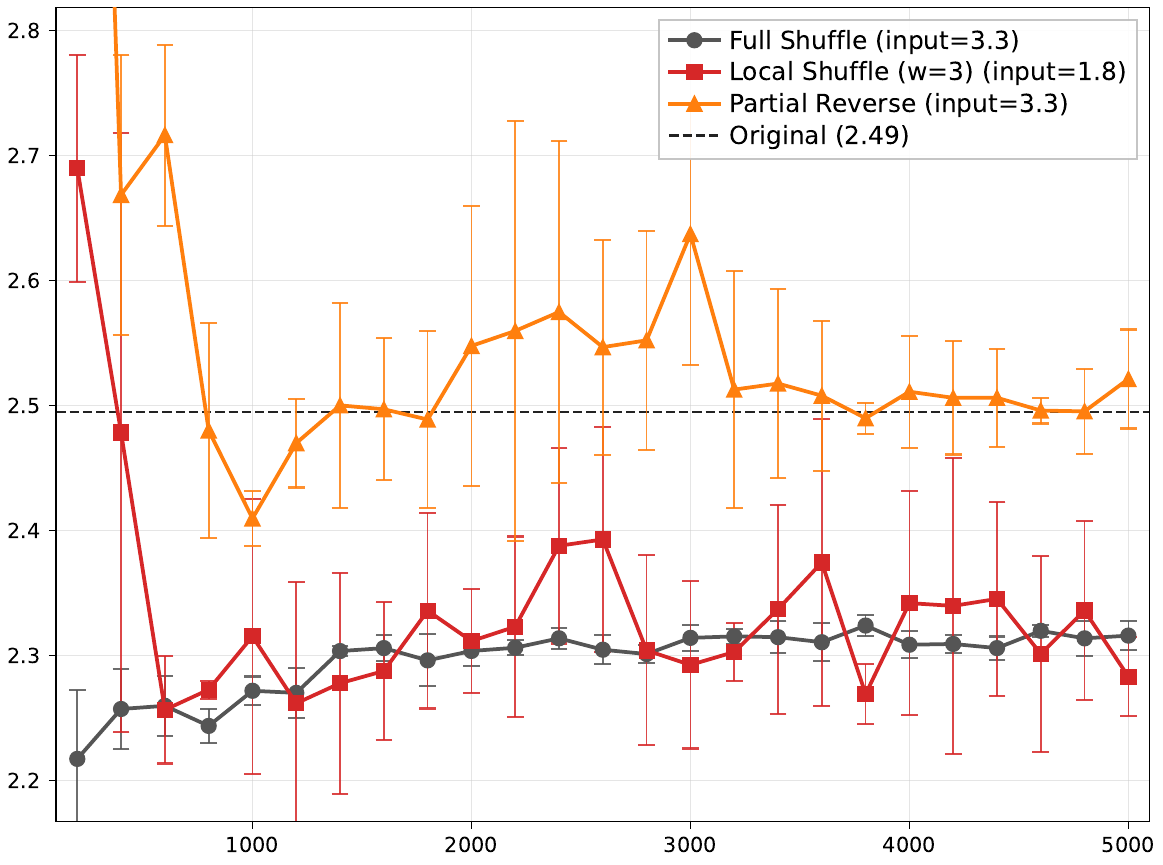}
  \caption{$ADL$ - \textsl{bnc\_spoken}}
  \label{fig:bnc:adl}
\end{subfigure}
\hfill
\begin{subfigure}{0.5\columnwidth}
  \centering
  \includegraphics[width=\linewidth]{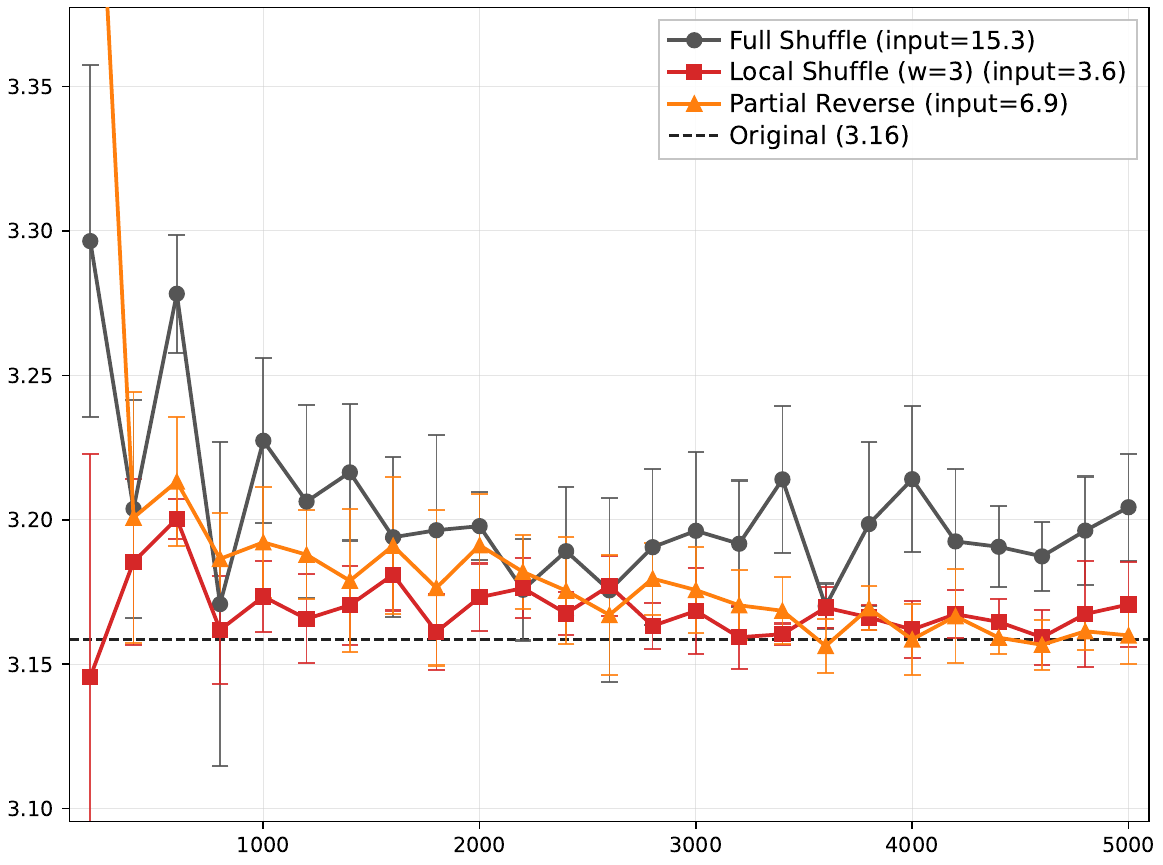}
  \caption{$ADL$ - \textsl{Gutenberg}}
  \label{fig:gutenberg:adl}
\end{subfigure}
\hfill
\begin{subfigure}{0.5\columnwidth}
  \centering
  \includegraphics[width=\linewidth]{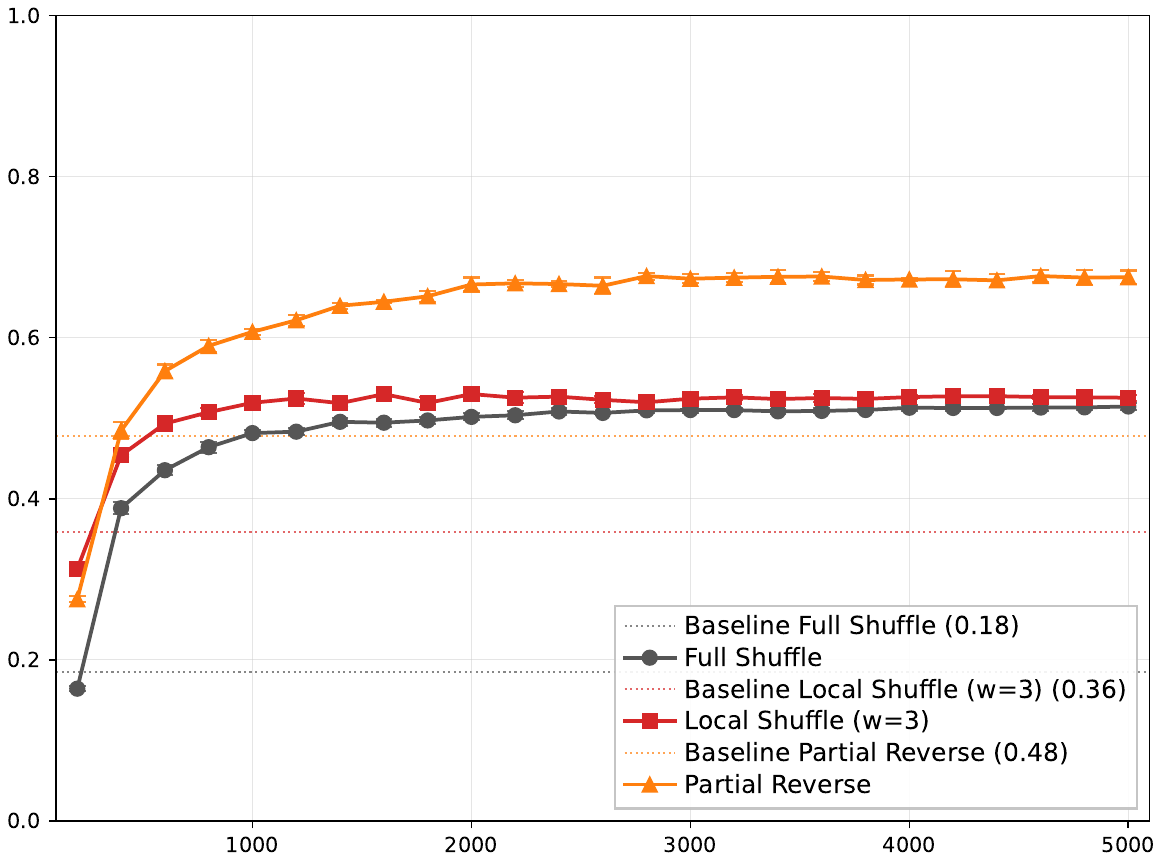}
    \caption{Triple $F_1$ - \textsl{bnc\_spoken}}
  \label{fig:bnc:f1}
\end{subfigure}
\hfill
\begin{subfigure}{0.5\columnwidth}
  \centering
  \includegraphics[width=\linewidth]{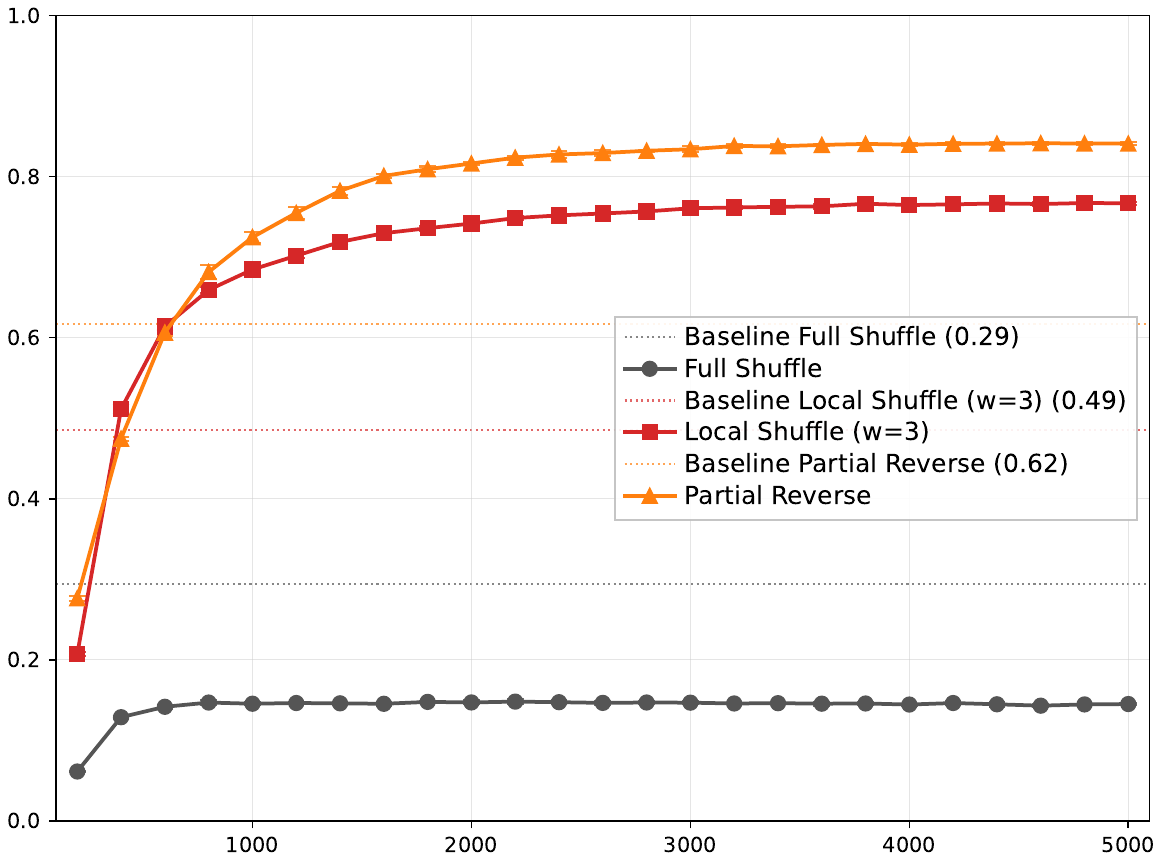}
    \caption{Triple $F_1$ - \textsl{Gutenberg}}
  \label{fig:gutenberg:f1}
\end{subfigure}
\caption{\footnotesize{Dependency-based evaluation across perturbation types and datasets. Dashed lines show per-perturbation scrambled-input baselines; solid line shows the natural language reference. (\subref{fig:bnc:adl}), (\subref{fig:gutenberg:adl}) \textsc{PartialReverse} and \textsc{LocalShuffle} converge toward natural ADL on both datasets; \textsc{FullShuffle} output ADL also approaches the natural value on \textsl{Gutenberg} despite near-zero EM, reflecting fluent but unfaithful generation (ADL is measured on the model's own output, not against the source). (\subref{fig:bnc:f1}), (\subref{fig:gutenberg:f1}) All three perturbation types surpass their scrambled-input baselines on \textsl{bnc\_spoken}; on \textsl{Gutenberg}, \textsc{PartialReverse} and \textsc{LocalShuffle} improve further while \textsc{FullShuffle} collapses.}}

\label{fig:dependencies}
\end{figure*}

\paragraph{Experiment 1:} 
Figure \ref{fig:bnc:em-bleu} shows learning curves across checkpoints on \textsl{bnc\_spoken}. The scrambled-input Dependency Triple F$_1$ baseline measures how much information locality survives each perturbation: \textsc{PartialReverse} (0.48) $>$ \textsc{LocalShuffle} (0.36) $>$ \textsc{FullShuffle} (0.18). \textsc{PartialReverse} improves most rapidly and reaches the highest final performance (EM $\approx$ 0.29, BLEU $\approx$ 0.55), while \textsc{LocalShuffle} and \textsc{FullShuffle} improve more slowly. On surface metrics, \textsc{FullShuffle} reaches a slightly higher EM ceiling ($\approx$ 0.23) than \textsc{LocalShuffle} ($\approx$ 0.17).

\noindent
Dependency Triple F$_1$ (Figures~\ref{fig:bnc:adl} and~\ref{fig:bnc:f1}) shows a consistent ordering throughout training - \textsc{PartialReverse} $\approx$ 0.67, \textsc{LocalShuffle} $\approx$ 0.53, \textsc{FullShuffle} $\approx$ 0.51 — indicating that EM underestimates \textsc{LocalShuffle}'s structural recovery. Under \textsc{LocalShuffle}, the model recovers more correct syntactic relationships than \textsc{FullShuffle} despite its lower EM, because exact match penalizes position errors regardless of structural correctness. \textsc{LocalShuffle}'s within-window scrambling produces errors that are locally bounded and structurally recoverable, while \textsc{FullShuffle}'s global randomisation creates dependency distances proportional to sentence length, limiting structural reconstruction. The ADL analysis (Figure~\ref{fig:bnc:adl}) further supports this: \textsc{PartialReverse} predictions converge most closely to the actual ADL ($\approx$2.49), while \textsc{LocalShuffle} predictions fall below baseline (predicted ADL $\approx$ 2.28). Qualitative examples at different checkpoints are provided in Appendix (Tables \ref{tab:app-example-shuffle}, \ref{tab:app-example-reverse}, and \ref{tab:app-example-hop}).

\paragraph{Experiment 2:}
Table \ref{tab:dataset-avg-len} shows average sentence lengths for both datasets. Comparing final performance (Figure \ref{fig:em-bleu}) reveals sentence length effects that diverge sharply across perturbation types. \textsc{PartialReverse} improves substantially from \textsl{bnc\_spoken} to \textsl{Gutenberg} (EM: 0.29 $\to$ 0.52; BLEU: 55\% $\to$ 79\%), and \textsc{LocalShuffle} also improves (EM: 0.17 $\to$ 0.25; BLEU: 40\% $\to$ 68\%). \textsc{FullShuffle}, however, collapses dramatically on longer sentences (EM: 0.23 $\to$ 0.01; BLEU: 42\% $\to$ 4\%), despite performing comparably to \textsc{PartialReverse} on short sentences (see also Appendix Figure \ref{fig:improvement}).

\noindent
This collapse is reflected in the ADL of the perturbed input: \textsc{FullShuffle} input ADL reaches 15.26 on \textsl{Gutenberg}, compared to 3.34 on \textsl{bnc\_spoken}, while \textsc{PartialReverse} and \textsc{LocalShuffle} show far more moderate increases (6.90 and 3.59 respectively). Dependency-based evaluation confirms this pattern: on \textsl{Gutenberg}, \textsc{LocalShuffle} and \textsc{PartialReverse} achieve F$_1$ scores of 0.77 and 0.84, compared to 0.53 and 0.67 on \textsl{bnc\_spoken}, while \textsc{FullShuffle} F$_1$ falls to 0.14 (Figures~\ref{fig:bnc:f1} and \ref{fig:gutenberg:f1}). ADL analysis (Figures~\ref{fig:bnc:adl} and \ref{fig:gutenberg:adl}) shows that \textsc{PartialReverse} and \textsc{LocalShuffle} predictions converge toward the actual ADL ($\approx$3.16) on \textsl{Gutenberg}; \textsc{FullShuffle} output ADL also sits near the natural value ($\approx$3.20) despite near-zero EM, but since ADL is measured on the model's own output, this reflects fluent generation rather than source recovery — confirmed by the collapse in Triple F$_1$, which is referenced against the gold structure.

\noindent

These results show that \textsc{PartialReverse}'s partial structure preservation and \textsc{LocalShuffle}'s bounded disruption enable recovery to scale with sentence length, whereas \textsc{FullShuffle}'s global shuffling creates dependency distances that grow proportionally with sentence length, exceeding the model's recovery capacity. Longer sequences increase dependency distances and enlarge the reordering search space, consistent with prior work on long-range dependency limitations \cite{Khandelwal2018Sharp,hahntheoretical2020,Tran2018Importance} and degraded order information under global disruption \cite{sinha2021masked,abdou2022word}.

\paragraph{Experiment 3:} 

We use the pre-trained (base) GPT-2 model — not the fine-tuned models from \citet{kallini2024mission} — to evaluate whether generated text aligns with general natural language distributions: lower perplexity indicates output more consistent with standard English.

\begin{figure}[tbp]
\centering
\begin{subfigure}{0.9\columnwidth}
  \centering
  \includegraphics[width=\linewidth]{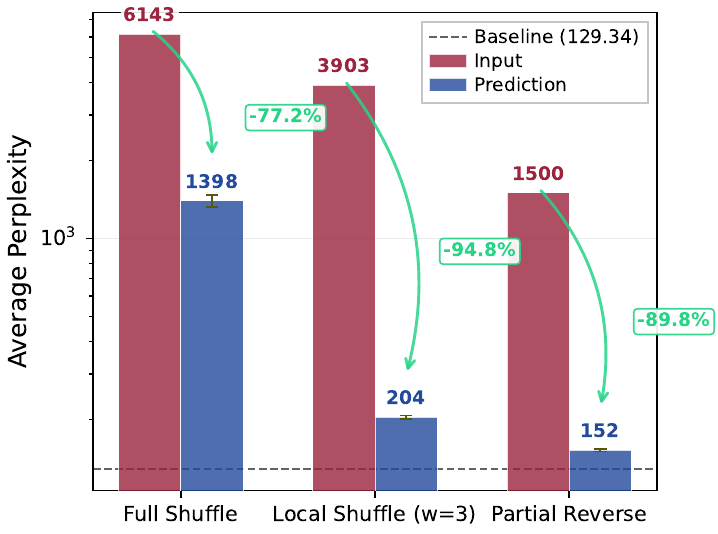}
\end{subfigure}
\caption{\footnotesize{Perplexity of perturbed inputs versus text predicted/recovered by each model, relative to baseline unperturbed text (129.34, dashed line), evaluated using pretrained GPT-2.}}
\label{fig:gutenberg-perplexity}
\end{figure}

\noindent
Figure \ref{fig:gutenberg-perplexity} compares perturbed input perplexity against model prediction perplexity, relative to the unperturbed text baseline (129.34). 
All three models show reduced prediction perplexity relative to their perturbed inputs.
\textsc{PartialReverse} input perplexity (1500.32) reduces to 152.38 (89.8\%), converging closest to baseline. \textsc{LocalShuffle} input perplexity (3902.56) reduces to 204.24 (94.8\%). \textsc{FullShuffle} shows the highest input perplexity (6143.02); predictions reduce to 1398.49 (77.2\%), yet remain far above baseline. The prediction perplexity ordering (\textsc{PartialReverse} $<$ \textsc{LocalShuffle} $\ll$ \textsc{FullShuffle}) mirrors the recovery-difficulty ordering seen across the other metrics. Concrete examples at different training stages are provided in Appendix Tables \ref{tab:app-example-shuffle}, \ref{tab:app-example-reverse}, and \ref{tab:app-example-hop}.

    \section{Discussion}
We investigate whether LLMs can recover accessible natural linguistic structure from systematically degraded input, and whether recovery difficulty scales with the degree of locality violation. Our results show that: recovery difficulty scales systematically with information locality violation; structural recovery measured by dependency metrics precedes surface reconstruction measured by EM; and sentence length interacts differently across perturbation types, amplifying recovery for \textsc{PartialReverse} and \textsc{LocalShuffle} while causing \textsc{FullShuffle} to collapse completely.

\paragraph{Information Locality.} Our results demonstrate that model performance reflects the degree to which perturbations violate information locality \citep{Gibson2000Dependency}. \textsc{PartialReverse} proved most tractable, as its preserved initial segment provides structural anchors for recovery. \textsc{LocalShuffle} and \textsc{FullShuffle} show comparable difficulty on short sentences, though the metric matters: \textsc{LocalShuffle} achieves higher Dependency Triple F$_1$ despite lower EM, indicating that surface reconstruction underestimates its structural recovery. This is because \textsc{LocalShuffle}'s within-window scrambling produces errors that are locally bounded and structurally recoverable, whereas \textsc{FullShuffle}'s global randomisation creates dependency distances proportional to sentence length, limiting structural reconstruction. \textsc{FullShuffle} collapses completely on longer texts for the same reason, while \textsc{LocalShuffle}'s bounded disruption allows recovery to scale. This ordering holds not only for final performance but also for the rate of improvement: \textsc{PartialReverse} converges rapidly, while \textsc{LocalShuffle} and \textsc{FullShuffle} improve more slowly, reflecting their greater degree of locality disruption.

\noindent
The Dependency Triple F$_1$ analysis reinforces this, showing that the model recovers a substantial portion of correct syntactic relationships even when full string reconstruction remains elusive. When we consider alignment with the natural-language distribution, for\textsc{LocalShuffle} and \textsc{PartialReverse}, prediction perplexity drops to near the unperturbed baseline and co-occurs with high EM and Triple F$_1$, so here low perplexity reflects genuinely recovered, locality-respecting text \citep{Hahn2020Universals,hahntheoretical2020}. For \textsc{FullShuffle}, perplexity also drops substantially, yet EM is near zero: the model produces fluent English without reconstructing the original. This dissociation shows that surface naturalness can be achieved without faithful recovery, and that accurate reconstruction must be established through the dependency-based metrics rather than perplexity alone.

\paragraph{Performance across perturbation functions.} 
\textsc{FullShuffle} shows intermediate performance on short sentences but collapses on longer ones, while \textsc{LocalShuffle} plateaus at a lower ceiling despite its narrower disruption window. This aligns with theoretical arguments about transformer limitations on unbounded hierarchical dependencies \citep{Tran2018Importance,deletang2023neural,hahntheoretical2020,Khandelwal2018Sharp}. 
The ADL analysis further reveals an architectural locality bias: under \textsc{LocalShuffle}'s within-window disruption, the model defaults to shorter-than-natural dependency arcs (predicted ADL $\approx$ 2.28 vs.\ actual 2.49 on \textsl{bnc\_spoken}). This pattern suggests that the model's  inductive bias prioritizes shorter dependencies, even when they are not always the correct ones, a tendency that is partially consistent with the principle of dependency length minimization observed in natural language \citep{Futrell2020Dependency}.

\noindent
The ordering of recovery difficulty across perturbations mirrors the learnability ordering reported by \citet{kallini2024mission} for the same perturbations. Performance varies gradually across the invertible conditions, while \textsc{FullShuffle} exhibits a clear categorical break, suggesting that the locality bias operates consistently regardless of the direction of learning.

\paragraph{Importance of sentence length.} Our second experiment revealed that sentence length interacts with perturbation type, reflecting information locality directly. In longer sentences, all perturbation types scatter related words further apart: \textsc{PartialReverse} reverses a longer trailing segment and \textsc{FullShuffle} disperses tokens across a wider range, both increasing absolute dependency distances. For \textsc{PartialReverse} and \textsc{LocalShuffle}, however, longer sentences also provide substantially richer training content — more co-occurring dependent elements that help the model learn which words belong together and how to restore their relationships. The performance improvement from \textsl{bnc\_spoken} to \textsl{Gutenberg} directly reflects this: when some local structure is preserved, the contextual benefit of longer training sentences outweighs the increased reordering difficulty. For \textsc{FullShuffle}, the opposite pattern emerges. Global randomization removes all structural anchors, causing dependency distances to grow beyond the point where contextual information can compensate, which leads to a complete collapse in performance. Longer sequences also exponentially increase the number of possible reorderings and integration costs as processing difficulty grows with dependency length \citep{Futrell2020Dependency,Gibson2000Dependency}. 
Although LLMs lack the cognitive and neurobiological constraints that drive information locality in humans, they exhibit architectural constraints that produce similar locality preferences: learning efficiency favors local dependencies, creating biases partially aligned with natural language properties \citep{Futrell2020Dependency,Khandelwal2018Sharp}.

\section{Conclusion}

This study demonstrates that LLMs can recover information from perturbed inputs, corresponding to strings of `linguistically impossible' languages. Recovery difficulty scales systematically with locality violation severity. 
Performance varies by perturbation type and sentence length — \textsc{PartialReverse} easiest, \textsc{FullShuffle} collapsing on longer sentences — dependency-based evaluation reveals substantial syntactic recovery even when exact reconstruction fails, and perplexity analysis confirms that recovered text is substantially more natural than the perturbed input. 
These results provide empirical support for information locality as a graded principle of efficient sequential processing whose effect on recovery parallels its reported effects on acquisition and processing. The ADL analysis further shows that the model's inductive bias favors shorter-than-natural dependency arcs — a structural alignment with dependency length minimization that emerges from attention architecture rather than explicit optimization. Ongoing work addresses three open questions: whether the recovery gradient observed at the output level reflects genuine internal representational change; whether syntactic encoding and causal use dissociate under recovery fine-tuning; and whether the locality gradient and the architecture's recovery capacity generalize across model families and scales.


\section{Limitations}

Our perturbations, inspired by previous work by \citet{kallini2024mission}, correspond to strings of `impossible languages' which are designed to violate linguistic universals. These represent only a small subset of the space of possible perturbations. Future work could explore a broader range of impossible language constructions, including those that violate other aspects of linguistic structure beyond word order, and evaluate whether consistent patterns emerge in model behavior. Additionally, investigating whether models can generalize their translation abilities to novel perturbation types or combinations would provide insight into whether they learn abstract principles of information locality recovery or task-specific mappings.

The notion of "impossible language" itself requires careful interpretation. While linguistic theory identifies certain systems as impossible for human acquisition based on violations of information locality and other neurobiological and cognitive constraints, these systems remain well-defined computational problems that neural networks can, in principle, solve. Our results highlight an important distinction between cognitive impossibility, which arises from violations of information locality, and computational possibility. Recognizing this gap is essential for understanding both the potential and the limits of LLMs as models of human language and as practical language processing tools.
    
\bibliography{custom}

@book{chomsky1965aspects,
  author    = {Noam Chomsky},
  title     = {Aspects of the Theory of Syntax},
  year      = {1965},
  publisher = {MIT Press},
  address   = {Cambridge, MA}
}

@book{Chomsky1986,
  author    = {Noam Chomsky},
  title     = {Knowledge of Language: Its Nature, Origin, and Use},
  year      = {1986},
  publisher = {Praeger},
  address   = {New York}
}

@book{Jackendoff2002,
  author    = {Ray Jackendoff},
  title     = {Foundations of Language: Brain, Meaning, Grammar, Evolution},
  year      = {2002},
  publisher = {Oxford University Press},
  address   = {Oxford}
}

@book{Pinker1994,
  author    = {Steven Pinker},
  title     = {The Language Instinct},
  year      = {1994},
  publisher = {William Morrow},
  address   = {New York}
}

@book{Saussure1916,
  author    = {Ferdinand de Saussure},
  title     = {Course in General Linguistics},
  year      = {1983},
  note      = {R. Harris (Trans.). Original work published 1916},
  publisher = {Duckworth},
  address   = {London}
}

@book{Hockett1958,
  author    = {Charles F. Hockett},
  title     = {A Course in Modern Linguistics},
  year      = {1958},
  publisher = {Macmillan},
  address   = {New York}
}

@book{Lyons1968,
  author    = {John Lyons},
  title     = {Introduction to Theoretical Linguistics},
  year      = {1968},
  publisher = {Cambridge University Press},
  address   = {Cambridge}
}

@book{Hawkins1994,
  author    = {John A. Hawkins},
  title     = {A Performance Theory of Order and Constituency},
  year      = {1994},
  publisher = {Cambridge University Press},
  address   = {Cambridge}
}

@book{Moro2016Impossible,
  author    = {Andrea Moro},
  title     = {Impossible Languages},
  year      = {2016},
  publisher = {MIT Press},
  address   = {Cambridge, MA}
}

@misc{WarstadtEtAl2023BabyLM,
  author       = {Alex Warstadt and Leshem Choshen and Aaron Mueller and Adina Williams and
                  Ethan Wilcox and Chengxu Zhuang},
  title        = {Call for Papers -- The BabyLM Challenge: Sample-efficient pretraining on a
                  developmentally plausible corpus},
  year         = {2023},
  eprint       = {2301.11796},
  archivePrefix= {arXiv},
  primaryClass = {cs.CL}
}

@article{RadfordEtAl2019GPT2,
  title={Language models are unsupervised multitask learners},
  author={Radford, Alec and Wu, Jeffrey and Child, Rewon and Luan, David and Amodei, Dario and Sutskever, Ilya and others},
  year         = {2019},
  howpublished = {OpenAI technical report}
}

@article{ZivEtAl2025BiaslessLLMs,
  author  = {Imry Ziv and Nur Lan and Emmanuel Chemla and Roni Katzir},
  title   = {Biasless Language Models Learn Unnaturally: How LLMs Fail to Distinguish the Possible from the Impossible},
  journal = {arXiv preprint arXiv:2510.07178},
  year    = {2025}
}

@book{chomsky1957syntactic,
  author    = {Noam Chomsky},
  title     = {Syntactic Structures},
  year      = {1957},
  publisher = {Mouton},
  address   = {The Hague}
}

@book{Nefdt2024,
  author    = {Ryan M. Nefdt},
  title     = {The Philosophy of Linguistics},
  year      = {2024},
  publisher = {Cambridge University Press},
  address   = {Cambridge}
}

@article{Smith1993,
  author  = {Neil V. Smith and Ianthi-Maria Tsimpli},
  title   = {The acquisition of possible and impossible languages by a polyglot savant},
  journal = {Lingua},
  year    = {1993},
  volume  = {91},
  number  = {1},
  pages   = {279--347}
}

@article{MussoEtAl2003BrocasArea,
  author  = {Mariacristina Musso and Andrea Moro and Volkmar Glauche and Michel Rijntjes and
             J{\"u}rgen Reichenbach and Christian B{\"u}chel and Cornelius Weiller},
  title   = {Broca's area and the language instinct},
  journal = {Nature Neuroscience},
  year    = {2003},
  volume  = {6},
  number  = {7},
  pages   = {774--781}
}

@inproceedings{mitchell2020priorless,
  author    = {Jennifer J. Mitchell and Jeffrey S. Bowers},
  title     = {Priorless Recurrent Networks Learn Curiously},
  booktitle = {Proceedings of the 28th International Conference on Computational Linguistics (COLING 2020)},
  year      = {2020},
  pages     = {5002--5015}
}

@inproceedings{huang2023lexinvariant,
  author    = {Qian Huang and Eric Zelikman and Sarah Li Chen and Yuhuai Wu and Gregory Valiant and Percy Liang},
  title     = {Lexinvariant Language Models},
  booktitle = {Advances in Neural Information Processing Systems},
  year      = {2023}
}

@inproceedings{abdou2022word,
  author    = {Mostafa Abdou and Vinit Ravishankar and Artur Kulmizev and Anders S{\o}gaard},
  title     = {Word Order Does Matter (and Shuffled Language Models Know It)},
  booktitle = {Proceedings of the 60th Annual Meeting of the Association for Computational Linguistics (ACL 2022)},
  year      = {2022},
  pages     = {6904--6919}
}

@inproceedings{sinha2021masked,
  author    = {Koustuv Sinha and Robin Jia and Dieuwke Hupkes and Joelle Pineau and Adina Williams and Douwe Kiela},
  title     = {Masked Language Modeling and the Distributional Hypothesis: Order Word Matters Pre-training for Little},
  booktitle = {Proceedings of the 2021 Conference on Empirical Methods in Natural Language Processing (EMNLP 2021)},
  year      = {2021},
  pages     = {2888--2913}
}

@article{Hahn2020Universals,
  author  = {Michael Hahn and Dan Jurafsky and Richard Futrell},
  title   = {Universals of word order reflect optimization of grammars for efficient communication},
  journal = {Proceedings of the National Academy of Sciences},
  year    = {2020},
  volume  = {117},
  number  = {5},
  pages   = {2347--2353}
}

@inproceedings{ebrahimi2020self,
  author    = {Javid Ebrahimi and Dhruv Gelda and Wei Zhang},
  title     = {How Can Self-Attention Networks Recognize Dyck-n Languages?},
  booktitle = {Findings of the Association for Computational Linguistics: EMNLP 2020},
  year      = {2020},
  pages     = {4307--4313}
}

@inproceedings{deletang2023neural,
  author    = {Gr{\'e}goire Del{\'e}tang and Anian Ruoss and Jordi Grau-Moya and Tim Genewein and Kevin Li Wenliang and Elliot Catt and Chris Cundy and Marcus Hutter and Shane Legg and Joel Veness and Pedro A. Ortega},
  title     = {Neural Networks and the Chomsky Hierarchy},
  booktitle = {Proceedings of the International Conference on Learning Representations (ICLR 2023)},
  year      = {2023}
}

@inproceedings{kallini2024mission,
  author    = {Julie Kallini and Isabel Papadimitriou and Richard Futrell and Kyle Mahowald and Christopher Potts},
  title     = {Mission: Impossible Language Models},
  booktitle = {Proceedings of the 62nd Annual Meeting of the Association for Computational Linguistics (ACL 2024)},
  year      = {2024},
  pages     = {14691--14714}
}

@article{Xu2025CanLM,
  title={Can Language Models Learn Typologically Implausible Languages?},
  author={Tianyang Xu and Tatsuki Kuribayashi and Yohei Oseki and Ryan Cotterell and Alexander Scott Warstadt},
  journal={ArXiv},
  year={2025},
  volume={abs/2502.12317},
  url={https://api.semanticscholar.org/CorpusID:276421906}
}

@inproceedings{Khandelwal2018Sharp,
  author    = {Urvashi Khandelwal and He He and Peng Qi and Dan Jurafsky},
  title     = {Sharp Nearby, Fuzzy Far Away: How Neural Language Models Use Context},
  booktitle = {Proceedings of ACL 2018},
  year      = {2018},
  pages     = {284--294},
  url       = {https://aclweb.org/anthology/P18-1027}
}

@inproceedings{Tran2018Importance,
  author    = {Ke Tran and Arianna Bisazza and Christof Monz},
  title     = {The Importance of Being Recurrent for Modeling Hierarchical Structure},
  booktitle = {Proceedings of EMNLP 2018},
  year      = {2018},
  pages     = {4731--4736},
  url       = {https://aclanthology.org/D18-1503}
}

@article{Gibson2000Dependency,
  title={The dependency locality theory: A distance-based theory of linguistic complexity},
  author={Gibson, Edward and others},
  journal={Image, language, brain},
  volume={2000},
  pages={95--126},
  year={2000}
}

@article{Futrell2020Dependency,
  author  = {Richard Futrell and Roger Levy and Edward Gibson},
  title   = {Dependency locality as an explanatory principle for word order},
  journal = {Language},
  year    = {2020},
  volume  = {96},
  number  = {2},
  pages   = {372--413}
}

@article{Gibson1998Storage,
  author  = {Edward Gibson},
  title   = {Linguistic complexity: locality of syntactic dependencies},
  journal = {Cognition},
  year    = {1998},
  volume  = {68},
  number  = {1},
  pages   = {1--76}
}

@article{Temperley2018Dependency,
  author  = {David Temperley},
  title   = {Dependency-Length Minimization and the Link between Word Order and Cognition},
  journal = {Cognitive Science},
  year    = {2018},
  volume  = {42},
  number  = {8},
  pages   = {3039--3061},
  doi     = {10.1111/cogs.12689}
}

@inproceedings{rajpurkar-etal-2016-squad,
    title = "{SQ}u{AD}: 100,000+ Questions for Machine Comprehension of Text",
    author = "Rajpurkar, Pranav  and
      Zhang, Jian  and
      Lopyrev, Konstantin  and
      Liang, Percy",
    editor = "Su, Jian  and
      Duh, Kevin  and
      Carreras, Xavier",
    booktitle = "Proceedings of the 2016 Conference on Empirical Methods in Natural Language Processing",
    month = nov,
    year = "2016",
    address = "Austin, Texas",
    publisher = "Association for Computational Linguistics",
    url = "https://aclanthology.org/D16-1264/",
    doi = "10.18653/v1/D16-1264",
    pages = "2383--2392"
}

@inproceedings{papineni-etal-2002-bleu,
    title = "{B}leu: a Method for Automatic Evaluation of Machine Translation",
    author = "Papineni, Kishore  and
      Roukos, Salim  and
      Ward, Todd  and
      Zhu, Wei-Jing",
    editor = "Isabelle, Pierre  and
      Charniak, Eugene  and
      Lin, Dekang",
    booktitle = "Proceedings of the 40th Annual Meeting of the Association for Computational Linguistics",
    month = jul,
    year = "2002",
    address = "Philadelphia, Pennsylvania, USA",
    publisher = "Association for Computational Linguistics",
    url = "https://aclanthology.org/P02-1040/",
    doi = "10.3115/1073083.1073135",
    pages = "311--318"
}

@article{JelinekEtAl1977Perplexity,
    author = {Jelinek, F. and Mercer, R. L. and Bahl, L. R. and Baker, J. K.},
    title = {Perplexity—a measure of the difficulty of speech recognition tasks},
    journal = {The Journal of the Acoustical Society of America},
    volume = {62},
    number = {S1},
    pages = {S63-S63},
    year = {2005},
    month = {08},
    issn = {0001-4966},
    doi = {10.1121/1.2016299},
    url = {https://doi.org/10.1121/1.2016299},
    eprint = {https://pubs.aip.org/asa/jasa/article-pdf/62/S1/S63/11558910/s63_5_online.pdf},
}

@article{Liu2008Dependency,
  author  = {Haitao Liu},
  title   = {Dependency distance as a metric of language comprehension difficulty},
  journal = {Journal of Cognitive Science},
  year    = {2008},
  volume  = {9},
  number  = {2},
  pages   = {159--191}
}

@article{hahntheoretical2020,
  author  = {Michael Hahn},
  title   = {Theoretical limitations of self-attention in neural sequence models},
  journal = {Transactions of the Association for Computational Linguistics},
  year    = {2020},
  volume  = {8},
  pages   = {156--171}
}

@inproceedings{buchholz2006conll,
  title={CoNLL-X shared task on multilingual dependency parsing},
  author={Buchholz, Sabine and Marsi, Erwin},
  booktitle={Proceedings of the tenth conference on computational natural language learning (CoNLL-X)},
  pages={149--164},
  year={2006}
}

@inproceedings{cai-knight-2013-smatch,
    title = "{S}match: an Evaluation Metric for Semantic Feature Structures",
    author = "Cai, Shu  and
      Knight, Kevin",
    editor = "Schuetze, Hinrich  and
      Fung, Pascale  and
      Poesio, Massimo",
    booktitle = "Proceedings of the 51st Annual Meeting of the Association for Computational Linguistics (Volume 2: Short Papers)",
    month = aug,
    year = "2013",
    address = "Sofia, Bulgaria",
    publisher = "Association for Computational Linguistics",
    url = "https://aclanthology.org/P13-2131/",
    pages = "748--752"
}

@article{ouyang2022training,
  title={Training language models to follow instructions with human feedback},
  author={Ouyang, Long and Wu, Jeffrey and Jiang, Xu and Almeida, Diogo and Wainwright, Carroll and Mishkin, Pamela and Zhang, Chong and Agarwal, Sandhini and Slama, Katarina and Ray, Alex and others},
  journal={Advances in neural information processing systems},
  volume={35},
  pages={27730--27744},
  year={2022}
}

@inproceedings{someya_information_2025,
    address = {Vienna, Austria},
    title = {Information {Locality} as an {Inductive} {Bias} for {Neural} {Language} {Models}},
    url = {https://aclanthology.org/2025.acl-long.1357},
    doi = {10.18653/v1/2025.acl-long.1357},
    language = {en},
    urldate = {2026-05-20},
    booktitle = {Proceedings of the 63rd {Annual} {Meeting} of the {Association} for {Computational} {Linguistics} ({Volume} 1: {Long} {Papers})},
    publisher = {Association for Computational Linguistics},
    author = {Someya, Taiga and Svete, Anej and DuSell, Brian and O’Donnell, Timothy J. and Giulianelli, Mario and Cotterell, Ryan},
    year = {2025},
    pages = {27995--28013},
}

@article{mollica_composition_2020,
	title = {Composition is the {Core} {Driver} of the {Language}-selective {Network}},
	volume = {1},
	issn = {2641-4368},
	doi = {10.1162/nol_a_00005},
	language = {eng},
	number = {1},
	journal = {Neurobiology of Language (Cambridge, Mass.)},
	author = {Mollica, Francis and Siegelman, Matthew and Diachek, Evgeniia and Piantadosi, Steven T. and Mineroff, Zachary and Futrell, Richard and Kean, Hope and Qian, Peng and Fedorenko, Evelina},
	year = {2020},
	pages = {104--134}
}

\appendix
    \definecolor{diffgreen}{RGB}{0,150,0}
\definecolor{diffred}{RGB}{200,0,0}

\newcommand{\simi}[1]{\textcolor{diffgreen}{\textbf{#1}}}
\newcommand{\dif}[1]{\textcolor{diffred}{\textbf{#1}}}

\section{Fine-tuning Implementation Details}
\label{sec:finetune_details}

All models were fine-tuned using Hugging Face Transformers (v4.35.0) with PyTorch 2.1.0, employing the masked instruction-following approach described in Section~\ref{sec:model-finetuning}.

    \textbf{Batch size:} Effective batch size of 512 (per-device 128 with 4 gradient accumulation steps).
    
    \textbf{Optimization:} AdamW optimizer with learning rate $2 \times 10^{-5}$, cosine schedule, 500 warmup steps, weight decay 0.01, and gradient clipping at norm 1.0.
    
    \textbf{Training steps:} 200 steps (10K dataset), 5,000 steps (100K dataset), determined by validation loss convergence.
    
    \textbf{Precision:} Mixed precision training with bfloat16 and TensorFloat-32 on NVIDIA H100 GPUs.
    
    \textbf{Regularization:} Label smoothing factor of 0.1.
    
    \textbf{Checkpointing:} Every 20 steps (10K) or 200 steps (100K). Best model selected by lowest evaluation loss on 10\% held-out validation set.
    
    \textbf{Reproducibility:} All fine-tuning experiments were repeated with five random seeds (1337, 2025, 314159, 42, 8675309).

\section{Examples}
During training, we evaluated model performance at multiple checkpoints to monitor learning progress. We tracked generated text quality across checkpoints for each impossible language type. This section provides examples illustrating the improvement trajectory throughout training (Tables \ref{tab:app-example-shuffle}, \ref{tab:app-example-reverse}, \ref{tab:app-example-hop}).

\begin{figure*}[ht]
\begin{subfigure}{\textwidth}
  \centering
  \includegraphics[width=0.75\linewidth]{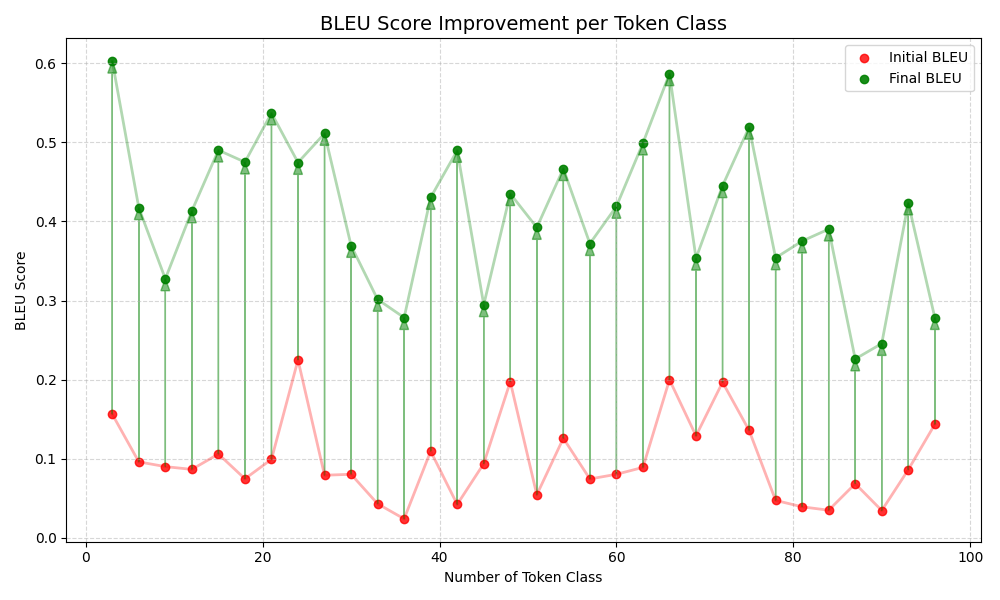}
\end{subfigure}
\caption{Improvement of the BLEU score for each class. The red line corresponds to the BLEU score before translating and for shuffled text, and the green line is for texts after translation by the final model.}
\label{fig:improvement}
\end{figure*}

Additionally, we investigate how performance varies by input text length for the \textsc{LocalShuffle} perturbation. Since \textsc{LocalShuffle} uses a window size of 3, we classify input texts by the number of shuffling windows they contain. Specifically, we define $\text{class}(text_i) = \lceil \text{len}(text_i) / 3 \rceil$, where $\text{len}(text_i)$ is the number of tokens in the text. This groups texts by how many 3-token windows were shuffled during perturbation. Figure \ref{fig:improvement} shows model improvement across these classes, revealing how performance scales with the number of shuffled segments.


\begin{table*}[ht]
\centering
\caption{Learning process for translation \textsc{LocalShuffle} at different learning checkpoints in a successful example. The \dif{red} colour determines the difference between the current text and with original text, and the \simi{green} colour determines similarity.}
\label{tab:app-example-shuffle}

\begin{tabularx}{\linewidth}{|X|c|}
\hline
\textbf{Input Text} &\textbf{Type}\\
\hline
Reggie was able to find out for them the place at which Fleming was putting up in Boston. Having ascertained from the clerk that he was still staying there, the next thing was to get hold of Louis Anderson. & original text \\
\hline
\dif{able} \simi{Reggie was} \simi{to find out} \dif{the }\simi{for them} \simi{place at which} \dif{putting} \simi{Fleming was} \dif{Boston in up ascert. Havingained the from that clerk he still was staying the there, next was thing hold} \simi{to get} \dif{Louis of Anderson .} & \textsc{LocalShuffle}\\
\hline
\dif{“Steve had} \simi{to find out} \dif{what they were going to put him in the hotel,} \simi{the next thing was} \dif{that was that he was going to put on to Tom. Having he was staying in the Boston car, which was the next thing} & Checkpoint-200\\
\hline
\simi{Reggie was able to find out for} \dif{the them} \simi{place at which Fleming was putting up in Boston. Having ascertained from the clerk that he was still staying there, the next thing was to get} \dif{on} \simi{Louis Anderson.} & Checkpoint-600\\
\hline
\simi{Reggie was able to find out for them the place at which Fleming was putting up in Boston. Having ascertained from the clerk that he was still staying there, the next thing was to get }\dif{to }\simi{Louis Anderson} \dif{of.} & Checkpoint-1000\\
\hline
\simi{Reggie was able to find out for them the place at which Fleming was putting up in Boston. Having ascertained from the clerk that he was still staying there, the next thing was to get hold of Louis Anderson.} & Checkpoint-1400\\
\hline
\end{tabularx}
\end{table*}

\begin{table*}[ht]
\centering
\caption{Learning process for translation \textsc{PartialReverse} at different learning checkpoints in a successful example. The \dif{red} colour determines the difference between the current text and with original text, and the \simi{green} colour determines similarity.}
\label{tab:app-example-reverse}

\begin{tabularx}{\linewidth}{|X|c|}
\hline
\textbf{Input Text} &\textbf{Type}\\
\hline
Bunny and Sue, so impatient they could hardly keep still, waited. They heard the front door open. They heard their father talking. Then came a funny, squeaking, whining sound. & original text \\
\hline
Bunny and Sue , so impatient \squared[white, colframe=black]{R} \dif{. sound whining , aking sque , funny a came Then . talking father their heard They . open door front the heard They . waited , still keep hardly could they} & \textsc{PartialReverse}\\
\hline
\dif{Bessie} \simi{and Sue, so impatient} \dif{and impatiently. The children were not allowed to keep them shut up. They were not allowed to keep them shut. They were not allowed to keep them shut, and they were not allowed to} & Checkpoint-200\\
\hline
\dif{Bucky} \simi{and Sue, so impatient they could hardly keep still,} \dif{waited, waited, waited, waited.} \simi{They heard their father talking. Then} \dif{Then} \simi{came a funny,} \dif{squeaky voice.} & Checkpoint-600\\
\hline
\dif{Bumper} \simi{and Sue, so impatient they could hardly keep still, waited. They heard the front door open. They heard their father talking. Then came a funny, squeaking, whining sound.} & Checkpoint-1000\\
\hline
\simi{Bunny and Sue, so impatient they could hardly keep still, waited. They heard the front door open. They heard their father talking. Then came a funny, squeaking, whining sound.} & Checkpoint-1400\\
\hline
\end{tabularx}
\end{table*}

\begin{table*}[ht]
\centering
\caption{Learning process for translation \textsc{FullShuffle} at different learning checkpoints in a successful example. The \dif{red} colour determines the difference between the current text and with original text, and the \simi{green} colour determines similarity.}
\label{tab:app-example-hop}

\begin{tabularx}{\linewidth}{|X|c|}
\hline
\textbf{Input Text} &\textbf{Type}\\
\hline
"If the papers in the packet are of the sort you think they are," he declared, "they will destroy them before they will permit you to get hold of them." & original text \\
\hline
 \dif{of, the " before packet" they are sort papersIf he the are you to in the destroy." of they themthey declared will think get permit hold you them," will}& \textsc{FullShuffle}\\
\hline
 \dif{The order} \simi{of the} \dif{envelope," they are all the papers," said the papers," said the papers. "They will be able to them to take them to them."} & Checkpoint-200\\
\hline
 \dif{of the before the packet," they are sort papers "If he are to the destroy} \simi{in the} \dif{of them." they declared them will get the} \simi{permit you} \dif{take them," will."}& Checkpoint-600\\
\hline
\dif{"The of before the " packet they" are sort papers he are the to destroy the destroy of they declared them."} \simi{they will} \dif{think you take hold them," will you will."}& Checkpoint-1000\\
\hline
\dif{"And, before the " packet they" sort papers are} \simi{the papers} \dif{he are to the destroy} \simi{in the} \dif{of they." declared them} \simi{they will} \dif{get} \simi{permit you} \dif{hold hold them," will} & Checkpoint-1400\\
\hline
\dif{"And, before the " packet they are sort papers are} \simi{the papers} \dif{he to destroy you} \simi{in the} \dif{destroy of them." they declared them will} \simi{get hold} \dif{you hold them," will} & Checkpoint-1800 \\
\hline
\dif{"And, before the packet they are sort papers "If he are the to destroy you} \simi{in the} \dif{destroy} \simi{of them.}\dif{" they declared them will get think you permit hold them," will} & Checkpoint-2200 \\ 
\hline
\dif{", before the packet they" packet papers are sort to the he are to you destroy} \dif{in the} \dif{of them." they declared them will get} \simi{get hold} \dif{them," will them,"} & Checkpoint-2600 \\
\hline
\dif{", before the " packet they are sort papers are} \simi{the papers} \dif{he are to you destroy} \simi{in the} \dif{of them." they declared them will} \simi{get hold} \dif{you hold them will,"} & Checkpoint-3000 \\
\hline
\dif{", before the " packet they are sort papers are the he if you are to destroy} \simi{in the} \simi{of them.}\dif{" they declared them will get think you permit hold them," will} & Checkpoint-3400 \\ 
\hline
\dif{", before the packet they are packet} \simi{the papers} \dif{if he are the to destroy you} \simi{in the} \dif{destroy of them." they declared them will get} \simi{get hold} \dif{you permit them will,"} & Checkpoint-3800 \\
\hline
\dif{"And of the packet before they are packet} \simi{the papers} \dif{if he are the to you to destroy} \simi{in the} \simi{of them.}\dif{" they declared them will get get you permit hold them," will} & Checkpoint-4200 \\ 
\hline
\dif{", before the packet they are packet} \simi{the papers} \dif{if he are the to destroy you} \simi{in the} \dif{destroy of them." they declared them will get} \simi{permit you} \dif{hold hold them," will} & Checkpoint-4600 \\
\hline
\dif{"  of the packet before they are packet} \simi{the papers} \dif{if he are the to you to destroy} \simi{in the} \dif{of them." they declared them will get} \simi{get hold} \dif{you permit them will,"} & Checkpoint-5000\\
\hline

\end{tabularx}
\end{table*}

\end{document}